\documentclass{article}
\usepackage{ijcai11}
\usepackage{amsmath}
\usepackage{hyperref}
\usepackage{float}
\usepackage{times}

\usepackage{algorithm}
\usepackage{algpseudocode}
\usepackage{algorithmicx}
\usepackage{listings}
\usepackage{graphicx}
\usepackage{caption}

\title{A complementary study on PlanGPT: Evaluation with defined Performance Metrics and comparison with a planner }
\author{Youssef Abdelkader, Humbert Fiorino, Damien Pellier  \\
Univ. Grenoble Alpes \-- LIG \\
Grenoble, France\\
{firstname.lastname@univ-grenoble-alpes.fr}} 

\begin{document}

\maketitle

\lstset{
basicstyle=\ttfamily,
breaklines=true
breakatwhitespace=true,
}

\begin{abstract}
  Automated Planning is a subfield of Artificial Intelligence (AI) where the main objective is generating a sequence of actions, known as a plan, that helps us reach a goal state from an initial state. A planning problem is defined by a set of objects, an initial state and a desired goal state. The objective is to compute a plan that'll lead us from the inital state to the goal state. Programs that generate plans are called planners.
  In this paper, we did a complementary study to the state-of-the-art LLM called PlanGPT which was released last year. We redid some experiments to verify whether planning with LLMs is \textbf{pertinent} and \textbf{worthwhile}. We also check whether the results obtained in the official PlanGPT paper for plan coverage were correct, and we also performed a more comprehensive study on PlanGPT's performance: in our paper PlanGPT's performance was evaluated using two metrics: Plan Cost and Plan Generation Time. The results of planGPT were compared to those produced by a traditional planner for the same plans and same metrics. We discovered that PlanGPT is no better than a Greedy search strategy.
\end{abstract}

\section{Introduction}
\subsection{Context}
Artificial Intelligence (AI) has taken the world by storm. A considerable amount of research is being conducted on AI to further try to improve upon it, to further understand it, to further optimize it, etc. In particular, we’re interested in two important notions that concern AI: Firstly, we will talk about Automated Planning which is a subfield of AI that focuses on generating an action sequence, called a plan, that enables robots to solve a planning problem. In this field, the Planning Domain Definition Language (PDDL) \cite{article} is used to define and formalize planning problems which are then fed to a planner, which is a program, to generate a solution. To formalize and define planning problems using PDDL, conventionally two PDDL files are created: A Domain file which defines the types (if the domain is typed) of objects, predicates, and actions that will be involved in the planning environment. A predicate is a condition that receives arguments (here in our case the arguments are objects) and checks whether the condition is true or false, so for instance "(iscar) car1" checks whether car1 is a car. An action, the founding block of the plan, has a name, parameters (an action receives objects), a pre-condition (a predicate that must be true before executing the action) and an effect (the predicate(s) that will be updated after executing the action). The other file is a Problem file which describes the objects that are involved in the plan (and optionally, their type), the initial state of the objects (where the objects are before doing the task), the goal state (where the objects should be after accomplishing the task). Secondly, we also will talk about Large Language Models (LLM), a new AI model that is used for Natural Language Processing (NLP) tasks; however, recently LLMs are used for other complex tasks such as image generation and text generation. \cite{bommasani2022opportunitiesrisksfoundationmodels}The bridge between these two notions is PlanGPT \cite{rossetti2024learning}, a LLM built from scratch, heavily inspired off of GPT-2, that is designed for planning. PlanGPT takes as input a planning problem and outputs a sequence of actions corresponding to the plan that solves the problem. PlanGPT tries to learn a general policy to solve a planning problem. Thus, PlanGPT is a LLM that plays the role of a planner. In the PlanGPT paper, after explaining what PlanGPT is and how it works, the authors test PlanGPT on the International Planning Competition (IPC) benchmark problems and show the plan coverage of PlanGPT on these problems. Afterwards, for the same problems, they compare PlanGPT to the state-of-the-art LLMs, Plansformer \cite{pallagani2022plansformergeneratingsymbolicplans} and the Graph Neural Network proposed by \cite{ståhlberg2022learninggeneraloptimalpolicies}, and show that PlanGPT has a better plan coverage. However, we don't know exactly which IPC benchmark problems were used for the comparisons. Another limitation is that they only show results for plan coverage, there are other interesting evaluation metrics that would've been pertinent for us to see such as Plan cost and Plan generation time, as these metrics would definitely help us in deciding whether planning with LLMs is \textbf{pertinent} and \textbf{worthwhile} or not. This is what we present in our paper, first we check whether the plan coverage results obtained here concord with those obtained in the PlanGPT paper, then we evaluate PlanGPT's performance using these two metrics on IPC benchmark planning problems and we even compare the results of PlanGPT with those of an optimal planner, called FastDownward \cite{Helmert_2006}, on the same planning problems and metrics. For the comparison we use two strategies: A* and Greedy. \\ The work is structured as follows: First, we give more details about PlanGPT and the modality of our experiment, then we present our results and finally say some concluding remarks. If you want to read more about planning with LLMs you may check the Related Work section or if you want to read about the challenges behind planning with LLMs, the purpose of this paper and the value that this paper may bring to the Automated Planning community (the motivation behind this paper), you may read the rest of the introduction.
\subsection{Related Work}
Quite a lot of research has been conducted to evaluate the planning capability of a LLM. An initial attempt was to prompt GPT-3.5 and GPT-4 to generate plans for planning problems \cite{valmeekam2022large} and unsurprisingly this yielded negative results. The next step was to try and customize LLMs for planning: a fine-tuned GPT2 for planning with a fine-tuned verifier \cite{arora2023learningleveragingverifiersimprove}  is one of the very first LLMs used for planning. It shows interesting results on a sample of planning domains, but also highlights that there is still a lot of room for improvement. Then, \cite{pallagani2022plansformergeneratingsymbolicplans} have implemented a newer model called Plansformer which is a fine-tuned LLM trained on a considerable amount of planning problems. Plansformer generates plans for a set of planning domains with varying complexities, demonstrating impressive adaptability and robustness. For a planning domain, the percentage of valid plans ranges from around 75 to 90\%.
Another fact to point out is that several other researchers have used other deep learning models for learning general policies. \cite{Toyer_2020} built a custom neural-network designed to generate plans in probabilistic planning. They use states and actions represented as neural layers and the goal is to obtain the next action in the plan thanks to the neural network. \cite{groshev2018learninggeneralizedreactivepolicies} built a Convolutional Neural Network (CNN) to solve planning problems in the Sokoban domain. This follows almost the same idea as \cite{Toyer_2020}, where a state in this case would be an image/screenshot of a Sokoban configuration. \cite{ståhlberg2022learninggeneraloptimalpolicies} have tried using a Graph Neural Network (GNN) to solve planning problems. In the GNN, they start with an initial state and the GNN computes all the reachable states from all the applicable actions. The GNN then selects the best state via a heuristic value. The main difference between these models and PlanGPT is that either they're fine-tuned and so not completely built for planning, or they are built from scratch however they focus more on state traversal and reachability whereas PlanGPT focuses more on generating the next valid action using a general policy. A closer approach would be the work of \cite{Serina2022APS} who trained a BERT model \cite{devlin2019bertpretrainingdeepbidirectional} from scratch with the aim of predicting missing actions from a partial plan. In our case, PlanGPT predicts the entire plan from scratch.
\subsection{Challenges and Ongoing Investigation}
Nonetheless, an ongoing investigation is being carried out when it comes to the planning capabilities of a LLM. We have seen that a LLM may be able to plan to some degree, but it is inconclusive, i.e,  there doesn’t seem to be a definite answer on whether a LLM can be a noteworthy planner and be able to generate correct plans for any planning problem, or whether planning for LLMs is simply too difficult for them and that they may be better suited for other tasks. The work of others has shown promising results highlighting the potential of this technology in this field, yet their results still show that there is a lot of room for improvement as no "general" LLM that works for any planning domain has been discovered yet. Even with efforts to advance the use of LLMs in this field, evaluating their planning capabilities is demanding as it is uncertain how much further improvement is realistically possible. And so, gauging a LLM's planning capabilities is extremely difficult.  This is why evaluating the planning capabilities of a LLM is a real challenge which requires further research.
\subsection{Statement of purpose}
 This paper will provide an in-depth analysis of the performance of the State-of-the-art LLM (PlanGPT) to determine whether planning with LLMs is possible and worthwhile, knowing that they necessitate a myriad of resources. Based on the research completed in this field, we hypothesize that as of now LLMs aren't capable of being competent planners.
 \subsection{Statement of value}
This paper may show to the Automated Planning community that leveraging LLMs for planning is a real possibility for the future, offering a compelling alternative to the traditional planners, with the potential to outperform them generally or in specific cases. It may also show that LLMs are not cut out for planning and that they may be well-suited for other tasks.

\section{PlanGPT details and Modality of Experiment}

\subsection{Overview}
In this section, we'll elaborate more on PlanGPT, i.e, explain what PlanGPT really is, how PlanGPT works, highlight some of PlanGPT's limits, etc, we will explain which metrics were used to define PlanGPT's performance which in turn supported our analysis, how the environment was configured and the procedure applied throughout.

\subsection{About PlanGPT}
As specified earlier, PlanGPT is a built-from-scratch LLM designed for planning. The goal is to help PlanGPT learn a per domain general policy, to apply in order to generate valid plans. Heavily inspired off of GPT-2, PlanGPT predicts the next token to generate a plan. \\
To understand what PlanGPT defines as a token, we must talk about the structure of a PDDL problem file and describe its contents. As mentioned earlier, there are three main components in a PDDL problem file:
\begin{itemize}
    \item 1: Objects section: Marked by ":objects", this section defines the involved objects in the plan. Objects may have types and the way to associate each object to a type depends on whether the domain is typed or not. For instance, if the domain is typed, one would write "car1 - vehicle" so to indicate that this car is of type vehicle. In an untyped version it'd be simply "car1" since there are no types.
    \item 2: Init section: Marked by ":init", this section defines the initial state of the problem. It is always a set of predicates and their associated objects. In this case, since we're in the initial state, all predicates defined are true. So for instance, "(on-table) BlockA" is a unary predicate which is true if and only if BlockA is on the table. Here, since we're in the inital state, this predicate is set to true.
    \item 3: Goal section: Marked by ":goal", it is also a set of predicates that must be valid at the end of the plan. If a plan is generated and these predicates are false then the plan is invalid.
\end{itemize}
These are the main sections. There are other parts such as a ":domain" section, which specifies the domain of the problem, however unlike the other sections PlanGPT does not tokenize this part. \\
PlanGPT tokenizes "fluents" which simply put are the predicates in the :init and :goal fields. A fluent corresponds to a predicate and its objects, so in the previous example, "(ontable) BlockA" would be a fluent. The way PlanGPT tokenizes these sections is by breaking down the fluent into separate words and then concatenating them all together. One word is one token. So for instance the following fluent "(on) BlockA BlockB" would have 3 three tokens: on, BlockA, BlockB. Then these tokens are concatenated to form: "On BlockA BlockB". This is now a sequence of tokens.
PlanGPT applies this operation for every fluent in the :init and :goal state. After it obtains all the sequences of tokens then the preprocessing phase is complete: this is what is fed into planGPT before predicting the next token. \\
After this preprocessing phase is complete, PlanGPT predicts the next token. However, in this case, PlanGPT does not predict predicates: It predicts actions and their associated objects. One action is a token and one object is another separate token. An action is the founding block of a plan, this is what allows progression in a plan. For instance, "(put-on) BlockA BlockB" would be an action which puts BlockA on Block B. In the context of plan generation, PlanGPT would generate: put-on, BlockA, BlockB (so 1 token for each word) and then add the end would concatenate everything. This continues like so until a "\textless end\textgreater" token is generated which symbolizes the end of the plan.\\
A dataset comprising of 70,000 planning problems was used to train PlanGPT. The procedure used is as follows: They use a problem generator to generate 70,000 planning problems, use LPG (a planner) to generate multiple valid plans per problem and then they randomize both the planning problem and the solution plans. Randomize in this context refers to assigning random names to objects, this is done to avoid bias since conventionally in PDDL problem files object names always end with a number to identify them (so for instance truck1, city1, car3...), this randomization step is done so that PlanGPT does not create "imaginary" links between the numbers in the objects, since object names should have no influence on the generated plan. \\
They also developed a new stopping technique called \textbf{Planning Coverage Early Stopping (CES)} in tandem with Cross Entropy which helps PlanGPT avoid overfitting. \\
Another very important detail is that PlanGPT \textbf{only works for 8 domains}. Those are: Blocksword, Satellite, Depot, Driverlog, Logistics, Zenotravel and Visitall. This is because every domain has an associated model implying that there are 8 models, which is a bit deceiving since one could argue that PlanGPT isn't 1 LLM but rather 8. This consumes a lot of resources. \\
You may now clearly see some of the limits of PlanGPT, it can only work on 8 domains, and within a problem file for a domain you may not give any arbitrary number of objects as there is a maximum limit. This caps the problem difficulty insinuating that highly complex problems are not runnable on PlanGPT. Another limitation is that nothing guarantees that the general policy learned by PlanGPT is optimal, which raises questions on the quality of generated plans. \\
To run PlanGPT with a set of PDDL problem files of a domain, a simple way is to write a per domain .json configuration file which contains pertinent information that helps PlanGPT generate plans. You can see those configuration files on the GitLab.
\subsection{PlanGPT Performance Metrics}
The goal of this paper is to provide an analysis of the performance of PlanGPT. However, before analysing the performance, we must somehow have a faithful and appropriate representation of the performance of PlanGPT. And so, we have decided that the following three metrics would define and represent the performance of PlanGPT across all domains:
\begin{itemize}
    \item Percentage of Plan Validation (Plan coverage): For every domain, we will run the International Planning Competition (IPC) benchmark PDDL problems on PlanGPT. We will ensure that these problems are compatible with PlanGPT thanks to domain-depended shell scripts that normalize the PDDL problem instances. The problems will be fed to PlanGPT and then the generated plans will be fed to a validator. PlanGPT comes with its own integrated validator, but for safety measure, we used the officially recognized validator VAL. The percentage of plan validation is pretty intuitive: It's the amount of correct plans over the total amount of plans.
    \item Plan Cost: For every domain, we will calculate the total amount of actions required to generate a plan for a problem, and we will do this for all generated plans of all the problems.
    \item Planning Time: For every domain, we will obtain (as this is already given by PlanGPT) the total amount of time required to generate a plan for a problem, and we will do this for all plans of all problems.
\end{itemize}

These are the metrics used to measure the performance of PlanGPT. We even decided to walk the extra mile and make our analysis stronger, not only will we measure PlanGPT's performance, we  compared it to traditional planners. This would really highlight how strong/weak PlanGPT is compared to traditional planners. This is why opted for FastDownward, an optimal planner, and to search for plans using two heuristics:
\begin{itemize}
    \item A*: This heuristic searches for the optimal plan cost for a planning problem, however it requires a lot of time doing so.
    \item FF Greedy: This heuristic outputs the first correct plan found. It is not optimal however it is very fast.
\end{itemize}
And so, our analysis involves two aspects: PlanGPT's performance across all domains using IPC Benchmark problems, and the comparison with the traditional planner FastDownward under the same circumstances and while using two search heuristics.
To properly view them, we drew one graph showing the plan cost and another for the plan generation time, across all problems for a domain. The graphs illustrated how each strategy performed across those problems. We will also compute the IPC scores for cost and generation time across all domains for all 3 strategies so as to have a global and comprehensive view of the results.

\subsection{Experimental setup}
Here we will talk about the configuration and organization of our environment so that our experiments may be replicated by others. In the Gitlab link provided above, you will find the following directories:
\begin{itemize}
    \item FastDownward: This folder contains the generated plans by the FastDownward Planner. Inside it you will have a folder for each heuristic used (A * and Greedy), and inside each folder you will find the plans generated for every domain (in the "SUCCESS" folder).
    \item PlansPlanGPT: This is the same idea as the previous folder except for the plans generated by PlanGPT. You will immediately find a folder per domain containing the successful plans that were generated, and the failed ones.
    \item planGPT: This is the planGPT code that is obtained from the Github repository. Inside the "scripts" directory, you will find all the models for all the domains (recall: one model per domain!), a CONFIGJSON directory which contains the per domain .json configuration files used to run planGPT and finally a "pddl\_files" folder which contains all the problem files (per domain) and the result .json file which contains all the generated plans and the planning generation time, per domain.
    \item Scripts: These are some scripts used to facilitate the process of plan generation.
    \item Plots: This folder contains all the plots, per domain, of the performance of PlanGPT, A * and Greedy. It also contains the tables for the IPC scores for time and cost.
\end{itemize}
Now that you understand the structure of our environment, we can proceed with its configuration. To run PlanGPT, we used the LIG Aker GPU cluster for faster computation. In the Gitlab we provide a documentation explaining how to access such infrastructures, and how to use them for running PlanGPT. If you don't have access to such infrastructures then fear not, PlanGPT does not require a strong GPU, based on our experience, a GPU NVIDIA GTX 1080 TI 11 GB was adequate for the job. Worst case you may even use your CPU. The server was configured with Python 3.11 and with some fundamental packages that were required to run PlanGPT (such as Tqdm, Datasets, Torch, Accelerate). The documentation talks about Conda however it is not necessary to use it. A virtual environment is recommended. Also ensure that both FastDownward and VAL are installed on your machine.

\subsection{Procedure}
Finally, in order to conduct our experiment, you must apply this procedure:
\begin{itemize}
    \item 1: Head to planGPT/scripts. This is where we will run planGPT. To run it, you must choose one of the 8 domains and after you have decided you must run it with the .json configuration file associated with the chosen domain. This will run all PDLL problem instances associated to that domain and output the solution on the screen. If you want to consult the results of PlanGPT you can head to: \textit{pddl\_Files/your\_domain/log\_generation/} and you should find a "LOGResults.json" file. An example command is provided in the Gitlab in the ReadMe.
    \item 2: Go to scripts and run the PlanGPT\_To\_ValidatorF script. It expects a results .json file (the one you just obtained in the previous step), a domain file (which is found in \textit{pddl\_files/your\_domain/domain} and a problem dir (which is found in \textit{pddl\_files/your\_domain/BENCHMARK\_PBS\_NORMALIZED} and an output directory. We recommend using the same naming convention established in PlansPlanGPT that is associated with your domain. You can find examples on how to run this script on the gitlab.
    \item 3: Now that you have obtained your planGPT results, it is time to do the same for FastDownward. Head to scripts and apply the "run\_and\_validateMulti.sh" which runs all PDDL problem instances of the domain you chose. This script expects a domain, the problems directory (so the same as in the previous step), a search heuristic and an output directory. Please also respect the naming convention in \textit{FastDownward/your\_heuristic/} that is associated with your domain. You can find examples on how to run this script on the gitlab. It is recommended to run this script twice, once using an A* heuristic and another using a greedy heuristic, on the same PDDL problem instances.
    \item 4: Now you can run in scripts "summarize\_costs.sh" which computes the cost of every successful plan generated and stores it in a file for a set of PDDL problem instances of a domain, for all 3 strategies. The script also copies the cost summary and time summary for all 3 strategies in the correct domain directory in the plots directory. An example command is given in the gitlab.
    \item 5: After having copied all costs and time summaries for PlanGPT, A* and Greedy, you may run the plt\_cost\_threeway script which takes as argument a cost summary for the A* strategy, Greedy and PlanGPT and then outputs a plot showing the performance of all 3 strategies across all PDDL problems for a domain. The same applies to the time summaries with the plt\_time\_threeway.
    \item 6: To obtain the IPC score tables, you must run table\_cost.py and table\_time.py. These give you the IPC scores for all problems per domain. Sadly the final table which gives us the IPC scores across all domains has been constructed manually.
With this you should be able to reproduce the same experiments as we did and obtain the same results, more or less.
\end{itemize}

\section{Results}

\subsection{Overview}
In this section we will present the obtained results. We would like to clarify that we ran our experiments on 7 domains, the one domain left out being Visitall. PlanGPT comes with many different parameters for plan generation. In our case, we simply generated one plan per problem, without sampling. Another important information that we'd like to clarify is how we computed the IPC scores \cite{IPC3SCORES}, it is actually two different formulas for cost and time:
\begin{itemize}
    \item Cost: The IPC score for cost is calculated as follows: Between all strategies, the one with the lowest cost for a problem is assigned the value 1, while the others are assigned a portion = $\frac{\text{lowest cost}}{\text{their cost}}$. We compute this for every problem and sum them up at the end for all problems for a domain.
    \item Time: The IPC score for time follows the formula:
    \[
    \min\left(1,\ 1 - \log\left(\frac{t}{T}\right)\right)
    \]
Where t is the time it took for plan generation and T is a timeout time = 300s.
\end{itemize}
For Blocksworld, Driverlog, Zenotravel, Floortile and Logistics, the benchmarks used can be found at the Downward Benchmark Repository.\footnote{\url{https://github.com/aibasel/downward-benchmarks}}
For Satellite and Depot, due to unrecognized predicates in the link above, we have used other benchmarks that can be retrieved from IPC-2002 domain page. \footnote{\url{https://ipc02.icaps-conference.org/domains.html}}

\subsection{Graphs}
We will now present some of the graphs obtained for plan cost and plan time, for all 3 strategies and for all domains. Below we show the cost graphs for Blocksworld and Driverlog:

\begin{figure}[H]
    \centering
    \includegraphics[width=1\linewidth, height = 0.7\linewidth]{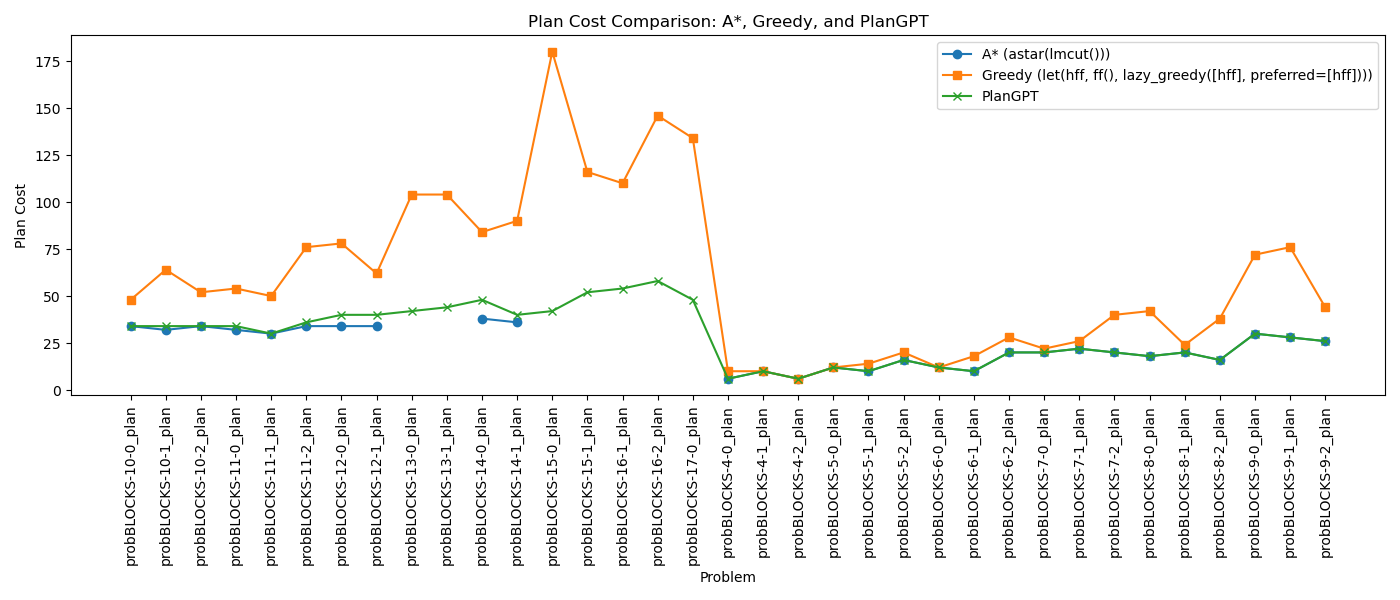}
    \caption{Plan cost across all problems for BlocksWorld domain, with PlanGPT in green, A* in blue and Greedy in Orange.}
    \label{fig:BW-cost}
\end{figure}

\begin{figure}[H]
    \centering
    \includegraphics[width=1\linewidth, height = 0.7\linewidth]{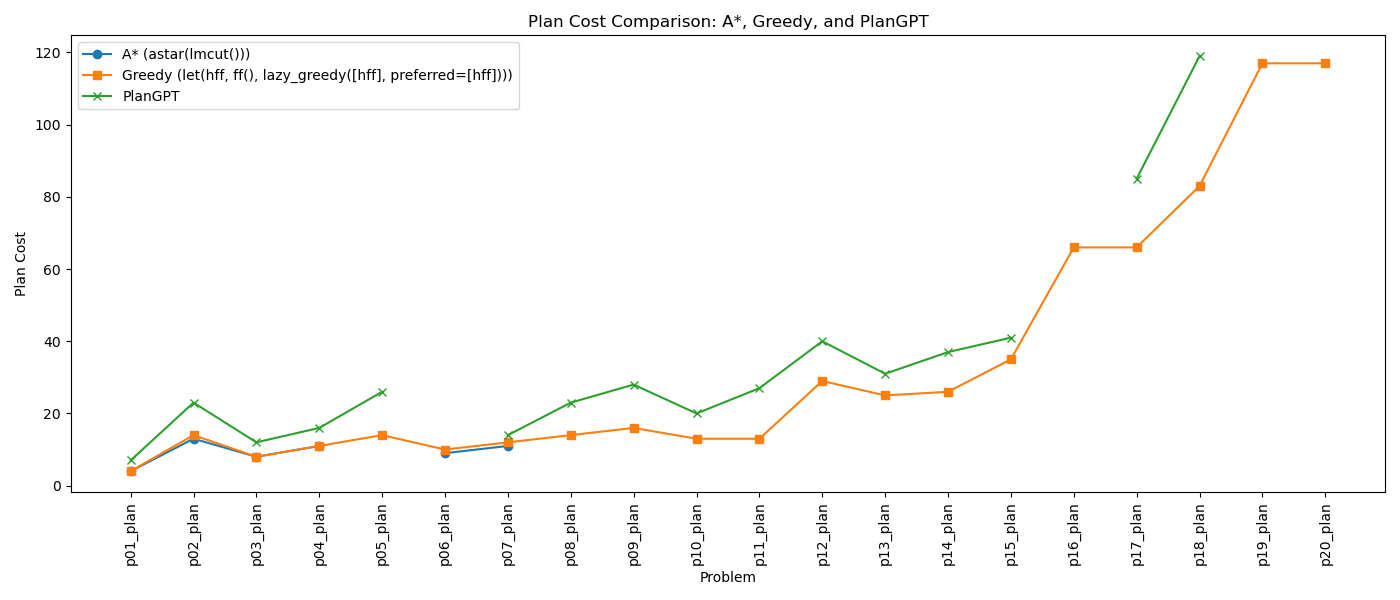}
    \caption{Plan cost across all problems for Driverlog domain with PlanGPT in green, A* in blue and Driverlog in orange.}
    \label{fig:DR-cost}
\end{figure}

In figure \ref{fig:BW-cost} we present the cost of the plans generated across all problems in the Blocksworld domain for all 3 strategies. First, we can see that at best PlanGPT matches the cost found by A* which is reasonable since A* always returns the optimal cost. Another important detail to note is that, in this case, PlanGPT always outperforms the greedy strategy as it always returns a lower (or equal) cost. In our case here the general policy applied to Blocksworld by PlanGPT is better than a Greedy search heuristic. Note that A* doesn't always have a plan cost, this is because A* timed out on that problem. However, in \ref{fig:DR-cost}, where we present the same cost metric but for the Driverlog domain, we have two issues: First, PlanGPT does not find a valid plan for all problems and second, when it does, it performs worse than a Greedy heuristic. This is to show that both ends of the spectrum are to consider: PlanGPT can completely outperform a Greedy strategy or the other way around. \\
Here we can present the time graphs:
\begin{figure}[H]
    \centering
    \includegraphics[width=1\linewidth, height = 0.7\linewidth]{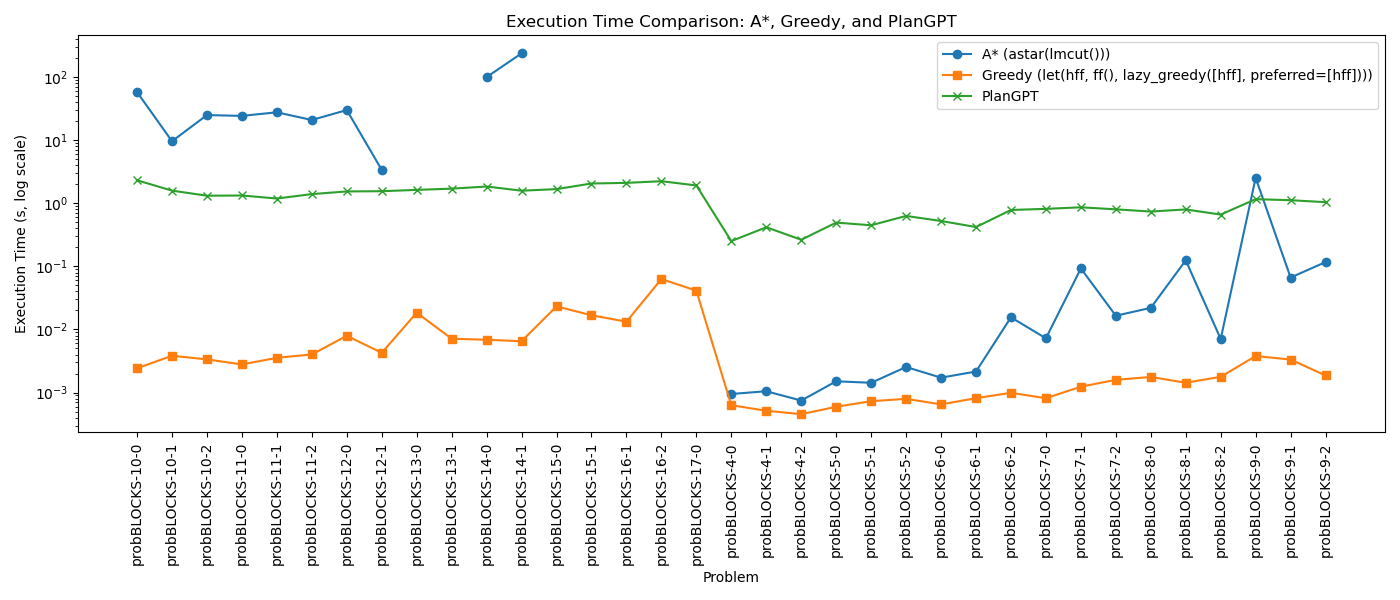}
    \caption{Planning time in \textbf{logarithmic scale} for all 3 strategies across all problems for the BlocksWorld domain, with same color assignment.}
    \label{fig:times-BW}
\end{figure}

\begin{figure}[H]
    \centering
    \includegraphics[width=1\linewidth, height = 0.7\linewidth]{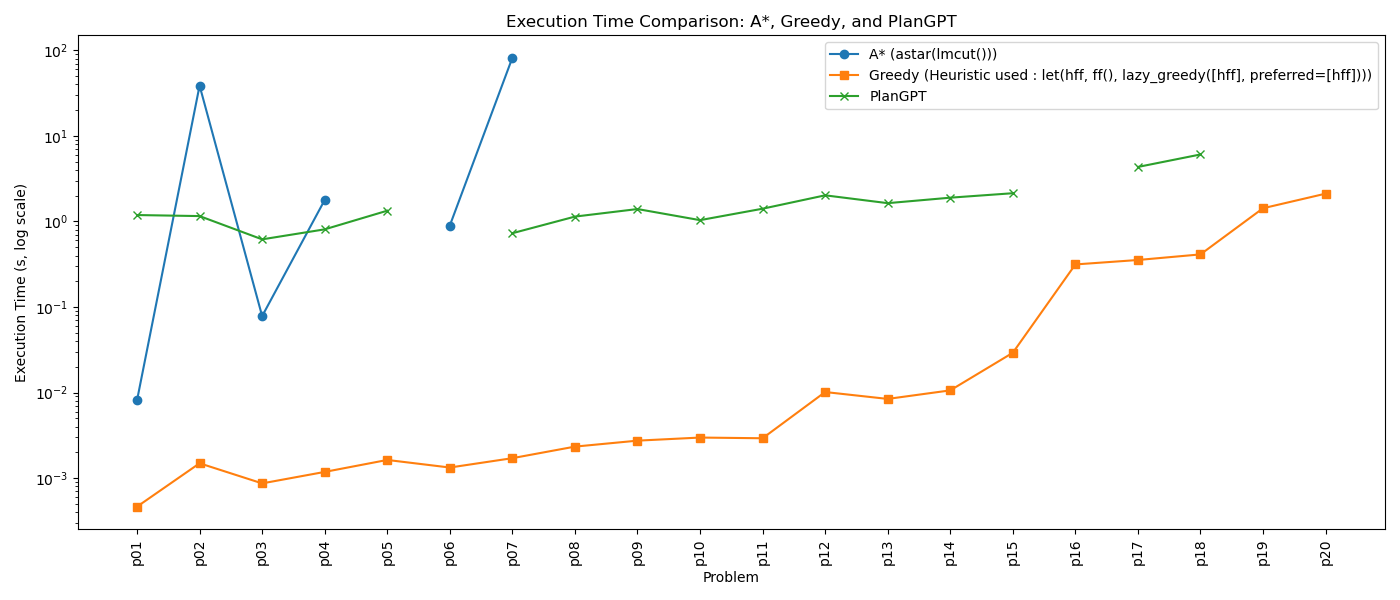}
    \caption{Planning time in \textbf{logarithmic scale} for all 3 strategies across all problems for the Driverlog domain, with same color assignment.}
    \label{fig:times-DR}
\end{figure}

In both \ref{fig:times-BW} and \ref{fig:times-DR}, we present the plan generation time across all problems for the 3 strategies for Blocksworld and Driverlog respectively. We see the same pattern: Greedy strategy seems to always be the fastest, with PlanGPT sometimes being faster than A* and other times not.
\subsection{IPC score Tables}
\begin{table}[ht]
\centering
\resizebox{1\linewidth}{!}{
\begin{tabular}{|c|c|c|c|c|c|c|}
\hline
Domain\\Statistics & \%PlanGPT & \%A* & \%Greedy  & IPC PlanGPT & IPC A* & IPC Greedy \\
\hline
BlocksWorld & 100\% & 80\% & 100\% & 0.97 & 0.8 & 0.6 \\
Depot & 85\% & 30\% & 80\% & 0.83 & 0.3 & 0.6 \\
Driverlog & 80\% & 30\% & 100\% & 0.53 & 0.3 & 0.99 \\
Logistics & 43\% & 71\% & 100\% & 0.37 & 0.71 & 0.99 \\
Satellite & 60\% & 41\% & 100\% & 0.56 & 0.41 & 0.95 \\
Zenotravel & 90\% & 61\% & 100\% & 0.78 & 0.67 & 0.93 \\
Floortile & 97\% & 7\% & 34\% & 0.96 & 0.07 & 0.3 \\
Total & 79\% & 46\% & 88\% & 5.02 & 3.26 & 5.36 \\
\hline
\end{tabular}
}
\caption{IPC \textbf{normalized} score computed across all tested domains for all 3 strategies for plan cost}
\label{tab:cost-table}
\end{table}

\begin{table}[ht]
\centering
\resizebox{1\linewidth}{!}{
\begin{tabular}{|c|c|c|c|c|c|c|}
\hline
Domain\\Statistics & \%PlanGPT & \%A* & \%Greedy  & IPC PlanGPT & IPC A* & IPC Greedy \\
\hline
BlocksWorld & 100\% & 80\% & 100\% & 0.95 & 0.6 & 1 \\
Depot & 85\% & 30\% & 80\% & 0.75 & 0.17 & 0.68 \\
Driverlog & 80\% & 30\% & 100\% & 0.73 & 0.22 & 0.99 \\
Logistics & 43\% & 71\% & 100\% & 0.4 & 0.55 & 1 \\
Satellite & 60\% & 41\% & 100\% & 0.56 & 0.35 & 1 \\
Zenotravel & 90\% & 61\% & 100\% & 0.80 & 0.51 & 1 \\
Floortile & 97\% & 7\% & 34\% & 0.75 & 0.02 & 0.15 \\
Total & 79\% & 46\% & 88\% & 4.95 & 2.46 & 5.82 \\
\hline
\end{tabular}
}
\caption{IPC \textbf{normalized} score computed across all tested domains for all 3 strategies for planning time}
\label{tab:time-table}
\end{table}

In \ref{tab:cost-table} and \ref{tab:time-table} we present the IPC scores of all 3 strategies across all domains. For the cost, overall Greedy is better, sometimes PlanGPT can outperform Greedy (such as in Blocksworld and Depot or even Floortile), but globally Greedy outperforms PlanGPT. For the time, aside from Floortile where PlanGPT has performed best and Driverlog, Greedy basically obtains the perfect score. As for the validation percentages, they're actually in-line with what \cite{rossetti2024learning} obtained in their paper since they also had tested PlanGPT on IPC Benchmark problems. The Plan Coverage is practically the same, there is on average around a 10\% difference in total with the results found in the official planGPT paper (here we use the IPC$^-$ set) and this slight difference further corroborates our findings for the IPC scores. (Note that the plan generation parameters are different).

\section{Discussion and Conclusion}
The goal of this study was to provide an analysis of PlanGPT's (the state-of-the-art) performance to determine whether planning with LLMs is possible, and if so, whether it is worth it. We had hypothesized that LLMs as of now aren't capable of being competent planners. Based on our findings, we can't exactly confirm nor deny our hypothesis: On one hand, PlanGPT's performance showed that for plan cost it was slightly worse than a Greedy search heuristic, and more specifically, it can be better in certain domains. If there is still room for improvement then there may be an LLM that completely outperforms the Greedy strategy and that isn't too bad on the plan generation time metric. On the other, knowing that PlanGPT has one model per domain, and acknowledging all of its limitations (fixed number of domain, fixed number of objects in total) it is still an interesting question to wonder, even if PlanGPT were to be improved and outperform Greedy, can consuming such an amount of resources for a better strategy than Greedy be considered "competent"? And is it even worth it? Not to mention that we only take into account the planning generation time for PlanGPT, we could've gone one step further and factored in training time in which case the performance would be much worse. Nonetheless, it is still interesting to wonder how PlanGPT can be improved even further while also limiting resource usage. The findings in this study may help other researchers gain some insight on the performance of the state-of-the-art LLM, and may help them ponder on how to improve the use of LLM in this field. Maybe even LLMs really aren't cut out for planning, but another model may be better suited? Such as Large Reasoning Models (LRM) proposed by  \cite{valmeekam2024llmscantplanlrms}.

\appendix


\end{document}